\documentclass[runningheads]{llncs}
\title{GMAT: Grounded Multi-Agent Clinical Description Generation for Text Encoder in Vision-Language MIL for Whole Slide Image Classification}

\usepackage{amsmath}

\usepackage[T1]{fontenc}
\usepackage{hyperref}
\usepackage{xcolor} 
\usepackage{multirow}
\usepackage{booktabs}  
\usepackage{amsmath, amssymb}
\usepackage{float}
 
\usepackage{graphicx,verbatim}
\begin{document}
%

%

\author{
    Ngoc Bui Lam Quang\inst{1}$^{\dagger}$\and
    Nam Le Nguyen Binh\inst{1}$^{\dagger}$\and
    Thanh-Huy Nguyen\inst{2}\and
    Le Thien Phuc Nguyen\inst{3}\and
    Quan Nguyen\inst{4}\and
    Ulas Bagci\inst{5} 
}

\authorrunning{Ngoc Bui Lam Quang, Nam Le Nguyen Binh, et al.}
%
\institute{AI VIETNAM, VietNam \and
Carnegie Mellon University, USA \and
University of Wisconsin-Madison \and
PTIT, VietNam \and
Northwestern University, USA\\
$^{\dagger}$Equal contribution.
}

\maketitle              
\begin{abstract}


Multiple Instance Learning (MIL) is the leading approach for whole slide image (WSI) classification, enabling efficient analysis of gigapixel pathology slides. Recent work has introduced vision-language models (VLMs) into MIL pipelines to incorporate medical knowledge through text-based class descriptions rather than simple class names. However, when these methods rely on large language models (LLMs) to generate clinical descriptions or use fixed-length prompts to represent complex pathology concepts, the limited token capacity of VLMs often constrains the expressiveness and richness of the encoded class information. Additionally, descriptions generated solely by LLMs may lack domain grounding and fine-grained medical specificity, leading to suboptimal alignment with visual features. To address these challenges, we propose a vision-language MIL framework with two key contributions: \textbf{(1)} A \textbf{grounded multi-agent description generation system} that leverages curated pathology textbooks and agent specialization (e.g., morphology, spatial context) to produce accurate and diverse clinical descriptions; \textbf{(2)} A \textbf{text encoding strategy using a list of descriptions} rather than a single prompt, capturing fine-grained and complementary clinical signals for better alignment with visual features. Integrated into a VLM-MIL pipeline, our approach shows improved performance over single-prompt class baselines and achieves results comparable to state-of-the-art models, as demonstrated on renal and lung cancer datasets.


\end{abstract}

\section{Introduction}
Pathological examination of tissue slides remains the gold standard for cancer diagnosis, offering high-resolution insights into cellular and structural abnormalities. However, whole slide images (WSIs) are gigapixel in size and contain complex, heterogeneous tissue patterns, making manual review labor-intensive and prone to variability.

To address the scale and complexity of WSIs, Multiple Instance Learning (MIL) has become the dominant approach for weakly supervised classification. In this framework, a slide is treated as a bag of image patches (instances), with supervision provided only at the slide level. Early models like ABMIL ~\cite{ilse2018attention-abmil}, CLAM ~\cite{lu2021data-clam}, and TransMIL ~\cite{shao2021transmil} leverage attention- or transformer-based aggregation to summarize patch features effectively. More recent work introduces advanced architectures, including HIPT ~\cite{chen2022scaling-hipt} with hierarchical transformers, DSMIL ~\cite{li2021dual-dsmil} with dual-stream learning, and others such as CAMIL~\cite{camil_fourkioti2024camil}, DTFD-MIL ~\cite{dtfd_zhang_dtfd-mil_2022}, SNUFFY~\cite{snuffy_afarinia2024snuffyefficientslideimage}, and DGMIL~\cite{qu2022dgmil}, each proposing novel strategies for aggregation or feature modeling to improve classification performance.

Vision-Language Models (VLMs) such as CLIP have been incorporated into MIL pipelines to improve classification and interpretability by aligning visual features with textual prompts like disease labels or descriptive phrases. A common strategy involves prompt tuning, using either handcrafted or LLM-generated prompts. MGPath ~\cite{nguyen2025mgpathvisionlanguagemodelmultigranular} introduces a multi-granular prompt framework that adapts across tissue magnifications to support few-shot learning. ViLaMIL ~\cite{nguyen2025fewshot} integrates magnification-aware embeddings with hierarchical attention across patch scales. MSCPT ~\cite{10979677} proposes a multi-scale, context-aware prompt tuning method that aligns vision-text embeddings across magnifications. These approaches highlight the potential of multiscale integration and prompt design in enhancing VLM-based WSI classification. These works underscore the role of prompt design and scale integration in advancing VLM-based digital pathology. Building on this, recent efforts are now leveraging foundation models to further extend vision-language and vision-only capabilities.

Recent progress in Whole Slide Imaging (WSI) has been driven by Vision-Language Models (VLMs) like BioCLIP~\cite{zhang2024biomedclip}, PLIP~\cite{huang2023plip}, and CONCH~\cite{lu2023mizero}, which align images with text for zero-shot performance. Dual-scale models such as ViLa-MIL~\cite{shi2024vilamil} enhance fine-grained reasoning. In parallel, vision-only models like CTransPath~\cite{wang2022ssltransformer} and GigaPath~\cite{xu2024slidefound} learn strong representations without text, enabling scalable classification in low-label settings.
 
Recent vision-language models (VLMs) have shown promising alignment between images and text, but typically rely on generic prompts that may not fully capture the clinical detail needed in computational pathology. This can make it challenging to represent the subtle, fine-grained patterns found in whole slide images (WSIs), especially in complex diagnostic settings. To support more clinically informed prompting, we propose the \textbf{Grounded Multi-Agent Text Generation (GMAT)} framework, which generates descriptive class texts using structured knowledge from pathology textbooks. At its core is \textbf{GMATG}, a lightweight, modular component that coordinates a set of simple, role-specific agents to guide the description process. While GMATG does not use advanced agent architectures, it provides a practical workflow for incorporating domain-specific information into text generation. \textbf{GMAT} integrates GMATG into a vision-language MIL pipeline to enhance both performance. In summary, our approach introduces two key innovations:
\begin{itemize}
\item A multi-agent system for generating clinically grounded descriptions, where each agent focuses on a distinct pathological attribute (e.g., cellular morphology, tissue architecture), enabling comprehensive and structured knowledge extraction;
\item A list-based text encoding strategy that replaces single-text prompts with multiple, diverse descriptions, capturing finer semantic details and improving alignment with visual features.
\end{itemize}

We integrate our approach into a vision-language MIL pipeline and demonstrate its effectiveness on renal and lung cancer datasets

\section{Methodology}
\textbf{Overview.} Our approach combines text-driven supervision with weakly supervised learning for WSI classification. It consists of two components: (1) \textbf{GMATG}, a multi-agent system that generates clinically grounded descriptions for each class, and (2) \textbf{GMAT}, a vision-language MIL model that aligns image patches with these descriptions using the CONCH encoder. Patch-level similarities are computed and aggregated using attention to produce slide-level predictions.
 
\subsection{Grounded Multi-Agent Text Generation (GMATG)}
\begin{figure}[h] 
\centering
\includegraphics[width=0.8\linewidth]{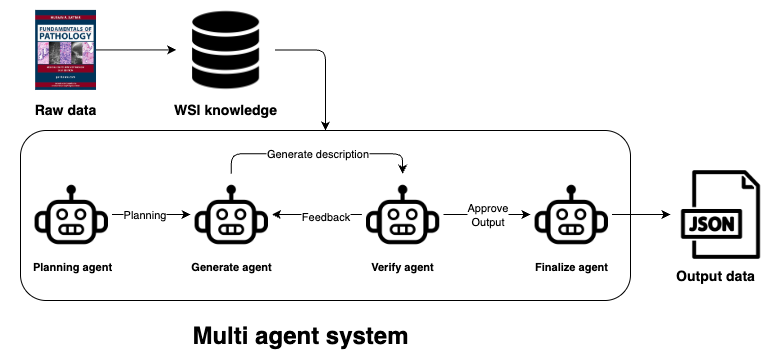}
\caption{Overview of GMATG. A team of specialized agents generates class-specific descriptions from domain knowledge, covering morphological, molecular, and clinical aspects. These are combined into rich text embeddings to guide visual understanding.}
\label{fig:gmatg}
\end{figure} 
To support class-specific description generation, we first build a structured knowledge base by extracting relevant disease-specific content from curated pathology textbooks, as shown in Figure ~\ref{fig:gmatg}. This serves as a shared foundation for all agents in the system, enabling consistent and medically grounded output.

The agents in this system, which could be powered by an advanced model like Gemini 2.5-Pro, perform distinct roles:
\begin{itemize}
\item \textbf{Planning Agent:} Creates a detailed guide for describing a specific cancer, outlining structure, rules for analyzing cell and tissue features, required clinical information, and quality standards. The output is a markdown plan with a summary, analysis instructions, and validation steps to guide the other agents. 
\item \textbf{Generate Agent:} Uses the plan and shared knowledge base to compose an initial draft of the class description.
\item \textbf{Verify Agent:} Reviews the written description for medical accuracy, completeness, and consistent terminology based on pathology standards. It produces a corrected version with a quality report and recommendations for improvement. 
\item \textbf{Finalize Agent:} Converts the approved description into a structured JSON file, using the cancer type as the main key and a list of short clinical sentences as values. These are ordered from general to microscopic, molecular, and clinical details. The agent ensures proper formatting, concise language, and removes all markdown.
\end{itemize}

Overall, this multi-agent workflow ensures structured planning, generation, and review of each class description. By combining expert-curated knowledge with specialized agent roles, GMATG generates clinically grounded, semantically rich prompts for downstream vision-language MIL classification.

\subsection{Vision-Language MIL Classification}  

\begin{figure}[h] 
    \centering
    \includegraphics[width=0.8\linewidth]{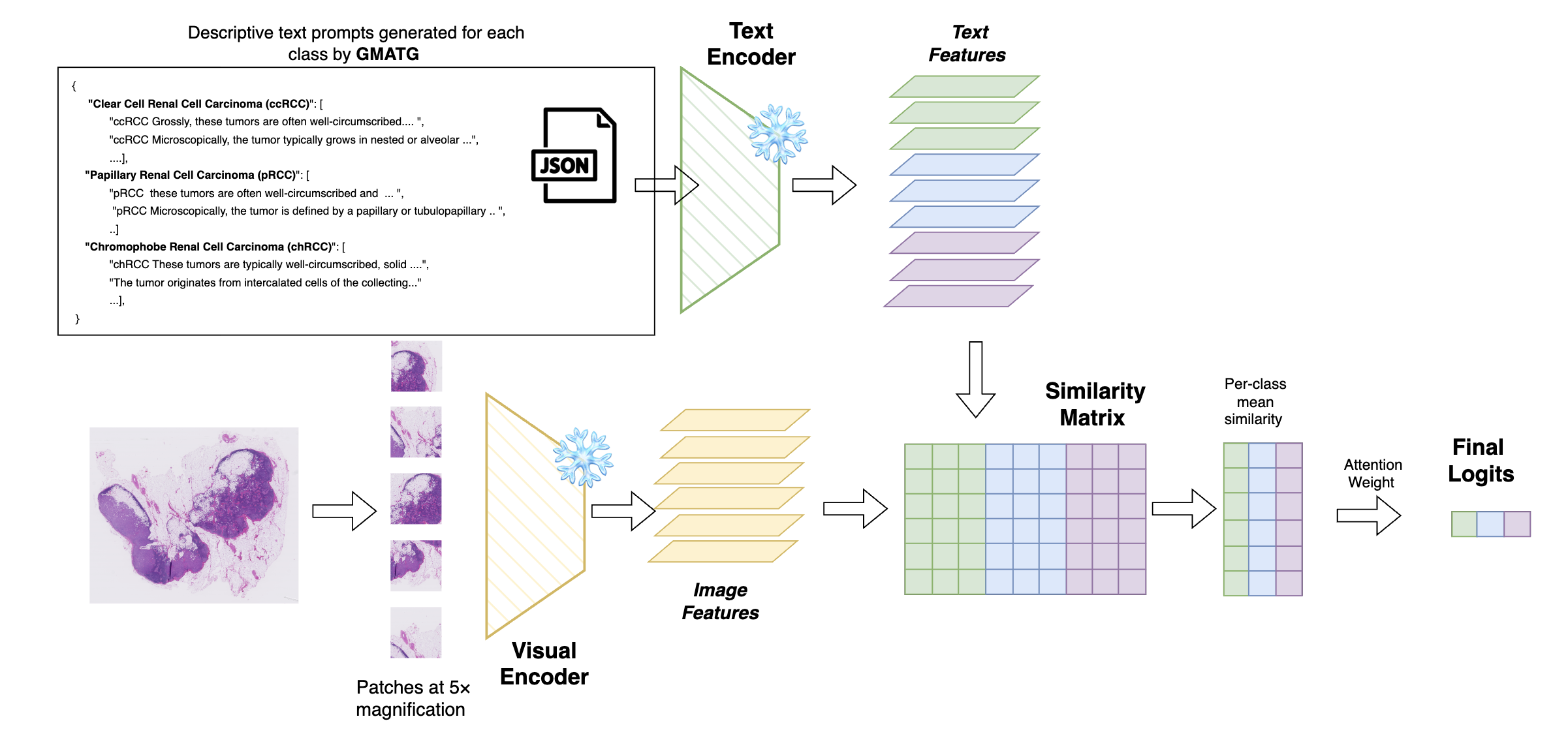}
    \caption{\textbf{Overview of our MIL framework for 5× magnification.} Patch features at 5× magnification are embedded using the CONCH visual encoder and matched with GMAT-generated text descriptions. Similarity scores are computed via visual-text alignment and aggregated using soft attention to produce class logits. Features from 10× magnification are processed in parallel, and predictions from both scales are fused for final slide-level classification.
    }
    \label{fig:mil_classification}
\end{figure} 


Our model, \textbf{GMAT}, uses a vision-language architecture with CONCH as a shared encoder for both image patches and text descriptions. Each whole slide image (WSI) is divided into patches at 5× and 10× magnification, which are processed by the CONCH visual encoder and mapped into a shared embedding space. For the text branch, we use multiple class-specific descriptions generated by GMATG. Stored in JSON format, these prompts capture diverse pathological features for each class. Each description is tokenized and encoded using the frozen CONCH text encoder to produce normalized text embeddings.

To align image and text, we compute the similarity between each patch embedding and all description embeddings. These per-patch similarity scores are aggregated into class-level scores by averaging over the descriptions corresponding to each class. Finally, we apply an attention-based aggregation mechanism, adapted from the CLAM model, to weight and combine patch-level class scores. This results in a slide-level prediction, trained using cross-entropy loss. The overall architecture is illustrated in \ref{fig:mil_classification}.

\section{Experiment}
\subsection{Datasets}
We evaluate on two cancer subtyping datasets: \textbf{TCGA-RCC (Renal)} with WSIs from Clear Cell (KIRC), Papillary (KIRP), and Chromophobe (KICH) subtypes, and \textbf{TCGA-Lung} with WSIs for Lung Adenocarcinoma (LUAD) and Lung Squamous Cell Carcinoma (LUSC). Both datasets use patient-level splits to prevent data leakage.
 
\subsection{Methods Under Comparision}
We evaluate zero-shot and fine-tuned performance across multiple settings. For zero-shot, we assess CONCH using either a single-description setup (denoted as \textit{Single Class Description} in the table) or a list of GMATG-generated descriptions. The single-description setup follows the two-level prompt format from ViLa-MIL~\cite{shi2024vilamil}, covering 5× and 10× magnifications. For fine-tuning, we compare ViLa-MIL with standard prompts to our GMAT framework. All methods use the same CONCH encoder for both text and image to ensure fairness. An ablation study further examines the impact of the multi-agent design versus a single-agent variant.

\subsection{Result and Analysis}
First, we evaluate the effectiveness of our approach in a zero-shot setting using the CONCH model. Specifically, we compare performance using a single class description versus a list of descriptions generated by GMATG. For the class description baseline, we adopt the two-level prompt structure used in ViLa-MIL ~\cite{shi2024vilamil}. In contrast, the list-based variant includes multiple, structured descriptions per class, covering diverse pathological aspects. Results show that the list-based approach consistently improves AUC, F1 score across both TCGA-RCC and TCGA-Lung datasets.
 
 


\begin{table}[H]
\centering
\footnotesize
\caption{\textbf{Zero-shot and Fine-tuned Performance on TCGA-RCC (Test Set)}\\
List-based descriptions are generated by the GMATG framework.}

\vspace{0.5em}

\begin{tabular}{|l|p{3.4cm}|c|c|c|}
\hline
\textbf{Model} & \textbf{Description Type} & \textbf{AUC (↑)} & \textbf{F1 Score (↑)} & \textbf{Accuracy (↑)} \\
\hline
\multicolumn{5}{|c|}{\textbf{Zero-shot Setting}} \\
\hline
CONCH & Single Class Description & 0.5730 {\scriptsize$\pm$ 0.0314} & 0.3466 {\scriptsize$\pm$ 0.0138} & 0.4821 {\scriptsize$\pm$ 0.0070} \\
CONCH & \textbf{List from GMATG} & \textbf{0.5912 {\scriptsize$\pm$ 0.0328}} & \textbf{0.3691 {\scriptsize$\pm$ 0.0347}} & 0.4357 {\scriptsize$\pm$ 0.0514} \\
\hline
\multicolumn{5}{|c|}{\textbf{Fine-tuned Setting}} \\
\hline
ViLa-MIL & Single Class Description & 0.9844 {\scriptsize$\pm$ 0.0070} & 0.9028 {\scriptsize$\pm$ 0.0445} & 0.9197 {\scriptsize$\pm$ 0.0184} \\
\textbf{GMAT} & \textbf{List from GMATG} & 0.9791 {\scriptsize$\pm$ 0.0116} & \textbf{0.9131 {\scriptsize$\pm$ 0.0293}} & \textbf{0.9262 {\scriptsize$\pm$ 0.0294}} \\
\hline
\end{tabular}\label{rcc}
\end{table}

\begin{table}[H]
\centering
\footnotesize
\caption{\textbf{Zero-shot and Fine-tuned Performance on TCGA-Lung (Test Set)}\\
List-based descriptions are generated by the GMATG framework.}

\vspace{0.5em}

\begin{tabular}{|l|p{3.4cm}|c|c|c|}
\hline
\textbf{Model} & \textbf{Description Type} & \textbf{AUC (↑)} & \textbf{F1 Score (↑)} & \textbf{Accuracy (↑)} \\
\hline
\multicolumn{5}{|c|}{\textbf{Zero-shot Setting}} \\
\hline
CONCH & Single Class Description & 0.6767 {\scriptsize$\pm$ 0.0288} & 0.6116 {\scriptsize$\pm$ 0.0164} & 0.6300 {\scriptsize$\pm$ 0.0089} \\
CONCH & \textbf{List from GMATG} & \textbf{0.7226 {\scriptsize$\pm$ 0.0233}} & \textbf{0.6693 {\scriptsize$\pm$ 0.0262}} & \textbf{0.6711 {\scriptsize$\pm$ 0.0247}} \\
\hline
\multicolumn{5}{|c|}{\textbf{Fine-tuned Setting}} \\
\hline
ViLa-MIL & Single Class Description & 0.9499 {\scriptsize$\pm$ 0.0308} & 0.8894 {\scriptsize$\pm$ 0.0418} & 0.8899 {\scriptsize$\pm$ 0.0422} \\
\textbf{GMAT} & \textbf{List from GMATG} & \textbf{0.9641 {\scriptsize$\pm$ 0.0057}} & \textbf{0.9023 {\scriptsize$\pm$ 0.0184}} & \textbf{0.9028 {\scriptsize$\pm$ 0.0183}} \\
\hline
\end{tabular}\label{lung}
\end{table}



We evaluate both zero-shot and fine-tuned performance on TCGA-RCC \ref{rcc} and TCGA-Lung \ref{lung}, comparing standard class-level descriptions with our list-based prompts generated by GMATG.

\textbf{Zero-shot.}
In the zero-shot setting, GMATG consistently improves performance over single-class descriptions. The improvements are especially notable on TCGA-Lung, where AUC rises from 0.6767 to 0.7226, along with gains in F1 and accuracy. This suggests that GMATG provides more informative and discriminative prompts, even without fine-tuning.

\textbf{Fine-tuned.}
With fine-tuning, GMAT achieves comparable performance ViLa-MIL on both datasets. On TCGA-RCC, it shows slight improvements in F1 and accuracy. On TCGA-Lung, GMAT achieves better results across all metrics, demonstrating the value of its structured, multi-agent descriptions during training.

Overall, GMAT achieves performance comparable to existing approaches in both zero-shot and fine-tuned settings, highlighting the potential of using richer, clinically grounded prompts. 
 
\subsection{Ablation} 
\begin{table}[h] 
\centering
\caption{\textbf{Ablation Study on TCGA-RCC}}
\vspace{0.3em}

\begin{tabular}{|l|c|c|c|}
\hline
\textbf{Model Variant} & \textbf{AUC (↑)} & \textbf{F1 Score (↑)} & \textbf{Accuracy (↑)} \\
\hline
Single Agent & 0.9776 ± 0.0109 & 0.9124 ± 0.0296 & 0.9239 ± 0.0304 \\
\textbf{Multi-Agent (GMATG)} & \textbf{0.9791 ± 0.0116} & \textbf{0.9131 ± 0.0293} & \textbf{0.9262 ± 0.0294} \\
\hline
\end{tabular}\label{abb-rcc}

\vspace{1em}

\caption{\textbf{Ablation Study on TCGA-Lung}}
\vspace{0.3em}

\begin{tabular}{|l|c|c|c|}
\hline
\textbf{Model Variant} & \textbf{AUC (↑)} & \textbf{F1 Score (↑)} & \textbf{Accuracy (↑)} \\
\hline
Single Agent & 0.9615 ± 0.0056 & 0.8968 ± 0.0145 & 0.8976 ± 0.0142 \\
\textbf{Multi-Agent (GMATG)} & \textbf{0.9641 ± 0.0057} & \textbf{0.9023 ± 0.0184} & \textbf{0.9028 ± 0.0183} \\
\hline
\end{tabular}\label{abb-lung} 
\end{table}
 
To evaluate the impact of the multi-agent design in GMATG, we compare it with a simpler single-agent version. In the single-agent setting, one agent extracts text from pathology textbooks and creates a list of class-specific descriptions without help or feedback from other agents. While it still uses a list-based approach, it lacks the structured, collaborative process of GMATG.

As shown in the ablation results for TCGA-RCC \ref{abb-rcc} and TCGA-Lung \ref{abb-lung}, the multi-agent system performs slightly better across all metrics. This suggests that having multiple agents working together leads to more accurate and comprehensive descriptions.


\section{Conclusion} 

We presented \textbf{GMAT}, a framework for vision-language MIL that generates clinically grounded, list-based prompts using a multi-agent system. By drawing from pathology textbooks, GMAT captures diverse and structured descriptions that align more effectively with visual features. Experiments on TCGA-RCC and TCGA-Lung show consistent improvements in both zero-shot and fine-tuned settings. These results highlight the value of domain-informed, collaborative prompt generation for enhancing performance and interpretability in computational pathology.
\section{Acknowledgement} 

We would like to thank AI VIETNAM for facilitating computational resources.

\bibliographystyle{ieeetr}  
\bibliography{ref.bib}  

@inproceedings{li2021dual-dsmil,
  title={Dual-stream multiple instance learning network for whole slide image classification with self-supervised contrastive learning},
  author={Li, Bin and Li, Yin and Eliceiri, Kevin W},
  booktitle={Proceedings of the IEEE/CVF conference on computer vision and pattern recognition},
  pages={14318--14328},
  year={2021}
}

@inproceedings{ilse2018attention-abmil,
  title={Attention-based deep multiple instance learning},
  author={Ilse, Maximilian and Tomczak, Jakub and Welling, Max},
  booktitle={International conference on machine learning},
  pages={2127--2136},
  year={2018},
  organization={PMLR}
}

@inproceedings{chen2022scaling-hipt,
  title={Scaling vision transformers to gigapixel images via hierarchical self-supervised learning},
  author={Chen, Richard J and Chen, Chengkuan and Li, Yicong and Chen, Tiffany Y and Trister, Andrew D and Krishnan, Rahul G and Mahmood, Faisal},
  booktitle={Proceedings of the IEEE/CVF conference on computer vision and pattern recognition},
  pages={16144--16155},
  year={2022}
}

@inproceedings{qu2022dgmil,
  title={Dgmil: Distribution guided multiple instance learning for whole slide image classification},
  author={Qu, Linhao and Luo, Xiaoyuan and Liu, Shaolei and Wang, Manning and Song, Zhijian},
  booktitle={International conference on medical image computing and computer-assisted intervention},
  pages={24--34},
  year={2022},
  organization={Springer}
}

@article{shao2021transmil,
  title={Transmil: Transformer based correlated multiple instance learning for whole slide image classification},
  author={Shao, Zhuchen and Bian, Hao and Chen, Yang and Wang, Yifeng and Zhang, Jian and Ji, Xiangyang and others},
  journal={Advances in neural information processing systems},
  volume={34},
  pages={2136--2147},
  year={2021}
}

@misc{snuffy_afarinia2024snuffyefficientslideimage,
    title={Snuffy: Efficient Whole Slide Image Classifier},
    author={Jafarinia, Hossein and Alipanah, Alireza and Razavi, Saeed and Mirzaie, Nahal and Rohban, Mohammad Hossein},
    booktitle={European Conference on Computer Vision},
    pages={243--260},
    year={2024},
    organization={Springer}
}

@inproceedings{camil_fourkioti2024camil,
  title={{CAMIL}: Context-Aware Multiple Instance Learning for Cancer Detection and Subtyping in Whole Slide Images},
  author={Olga Fourkioti and Matt {De Vries} and Chris Bakal},
  booktitle={The Twelfth International Conference on Learning Representations},
  year={2024},
  url={https://openreview.net/forum?id=rzBskAEmoc}
}

@misc{dtfd_zhang_dtfd-mil_2022,
	title = {{DTFD}-{MIL}: Double-Tier Feature Distillation Multiple Instance Learning for Histopathology Whole Slide Image Classification},
	url = {http://arxiv.org/abs/2203.12081},
	doi = {10.48550/arXiv.2203.12081},
	shorttitle = {{DTFD}-{MIL}},
	number = {{arXiv}:2203.12081},
	publisher = {{arXiv}},
	author = {Zhang, Hongrun and Meng, Yanda and Zhao, Yitian and Qiao, Yihong and Yang, Xiaoyun and Coupland, Sarah E. and Zheng, Yalin},
	urldate = {2024-12-18},
	date = {2022-03-22},
	eprinttype = {arxiv},
	eprint = {2203.12081 [cs]},
}

@misc{nguyen2025mgpathvisionlanguagemodelmultigranular,
      title={MGPATH: Vision-Language Model with Multi-Granular Prompt Learning for Few-Shot WSI Classification}, 
      author={Anh-Tien Nguyen and Duy Minh Ho Nguyen and Nghiem Tuong Diep and Trung Quoc Nguyen and Nhat Ho and Jacqueline Michelle Metsch and Miriam Cindy Maurer and Daniel Sonntag and Hanibal Bohnenberger and Anne-Christin Hauschild},
      year={2025},
      eprint={2502.07409},
      archivePrefix={arXiv},
      primaryClass={cs.CV},
      url={https://arxiv.org/abs/2502.07409}, 
}

@inproceedings{
nguyen2025fewshot,
title={Few-Shot Whole Slide Pathology Classification with Multi-Granular Vision-Language Models},
author={Anh-Tien Nguyen and Duy Minh Ho Nguyen and Nghiem Tuong Diep and Trung Quoc Nguyen and Nhat Ho and Jacqueline Michelle Metsch and Miriam Cindy Maurer and Daniel Sonntag and Hanibal Bohnenberger and Anne-Christin Hauschild},
booktitle={ICLR 2025 Workshop on Foundation Models in the Wild},
year={2025},
url={https://openreview.net/forum?id=nJZtYrOeoV}
}

@article{lu2021data-clam,
  title={Data-efficient and weakly supervised computational pathology on whole-slide images},
  author={Lu, Ming Y and Williamson, Drew FK and Chen, Tiffany Y and Chen, Richard J and Barbieri, Matteo and Mahmood, Faisal},
  journal={Nature biomedical engineering},
  volume={5},
  number={6},
  pages={555--570},
  year={2021},
  publisher={Nature Publishing Group UK London}
}

@ARTICLE{10979677,
  author={Han, Minghao and Qu, Linhao and Yang, Dingkang and Zhang, Xukun and Wang, Xiaoying and Zhang, Lihua},
  journal={IEEE Transactions on Medical Imaging}, 
  title={MSCPT: Few-shot Whole Slide Image Classification with Multi-scale and Context-focused Prompt Tuning}, 
  year={2025},
  volume={},
  number={},
  pages={1-1},
  doi={10.1109/TMI.2025.3564976}}

@article{zhang2024biomedclip,
  title={BiomedCLIP: A Multimodal Biomedical Foundation Model Pretrained from Fifteen Million Scientific Image-Text Pairs},
  author={Zhang, Sheng and Xu, Yanbo and Usuyama, Naoto and Xu, Hanwen and Bagga, Jaspreet and Tinn, Robert and Preston, Sam and Rao, Rajesh and Wei, Mu and Valluri, Naveen and others},
  year={2024},
  journal={arXiv preprint arXiv:2403.xxxxx}
}

@article{huang2023plip,
  title={A visual--language foundation model for pathology image analysis using medical twitter},
  author={Huang, Zhi and Bianchi, Federico and Yuksekgonul, Mert and Montine, Thomas J and Zou, James},
  journal={Nature Medicine},
  volume={29},
  number={9},
  pages={2307--2316},
  year={2023}
}

@inproceedings{lu2023mizero,
  title={Visual language pretrained multiple instance zero-shot transfer for histopathology images},
  author={Lu, Ming Y and Chen, Bowen and Zhang, Andrew and Williamson, Drew FK and Chen, Richard J and Ding, Tong and Le, Long Phi and Chuang, Yung-Sung and Mahmood, Faisal},
  booktitle={Proceedings of the IEEE/CVF Conference on Computer Vision and Pattern Recognition},
  pages={19764--19775},
  year={2023}
}

@inproceedings{shi2024vilamil,
  title={ViLa-MIL: Dual-scale vision-language multiple instance learning for whole slide image classification},
  author={Shi, Jiangbo and Li, Chen and Gong, Tieliang and Zheng, Yefeng and Fu, Huazhu},
  booktitle={Proceedings of the IEEE/CVF Conference on Computer Vision and Pattern Recognition},
  pages={11248--11258},
  year={2024}
}

@article{wang2022ssltransformer,
  title={Transformer-based unsupervised contrastive learning for histopathological image classification},
  author={Wang, Xiyue and Yang, Sen and Zhang, Jun and Wang, Minghui and Zhang, Jing and Yang, Wei and Huang, Junzhou and Han, Xiao},
  journal={Medical Image Analysis},
  volume={81},
  pages={102559},
  year={2022}
}

@article{xu2024slidefound,
  title={A whole-slide foundation model for digital pathology from real-world data},
  author={Xu, Hanwen and Usuyama, Naoto and Bagga, Jaspreet and Zhang, Sheng and Rao, Rajesh and Naumann, Tristan and Wong, Cliff and Gero, Zelalem and Gonzalez, Javier and Gu, Yu and others},
  journal={Nature},
  pages={1--8},
  year={2024}
}

\end{document}